\pdfoutput=1
\PassOptionsToPackage{table}{xcolor}
\documentclass[11pt]{article}

\usepackage{ACL2023}

\usepackage{times}
\usepackage{latexsym}

\usepackage[table]{xcolor}
\usepackage{graphicx}
\usepackage{graphics}
\usepackage{CJK}
\usepackage{tabularx}
\usepackage{bm}
\usepackage{amssymb}
\usepackage{booktabs}
\usepackage{url}
\usepackage{multicol}
\usepackage{tabularx}
\usepackage{booktabs}
\usepackage{color}
\usepackage{arydshln}
\usepackage{multirow}
\usepackage{amsmath} 
\usepackage{bm}
\usepackage{subfigure}
\usepackage{amstext}
\usepackage{tikz}
\usepackage{pgfplots}
\usepackage{hyperref}
\usepackage{subfigure}
\usepackage{enumitem}
\usepackage{mathtools}

\usepackage[T1]{fontenc}

\usepackage[utf8]{inputenc}

\usepackage{microtype}

\usepackage{inconsolata}

\usepackage{soul}

\usepackage{hyperref}

\usepackage{comment}
\usepackage{multirow}
\newcommand{\eg}{e.g., }

\newcommand{\Flodial}{\textit{FloDial}}
\definecolor{dgreen}{rgb}{0,0.55,0}
\definecolor{mgreen}{rgb}{0,0.7,0}

\newcommand{\myitem}{\vspace*{-1mm}\item}
\newcommand{\pvalue}{\mbox{\it p-value}}

\usepackage{amsmath,amsfonts,bm}

\def\eqref#1{equation~\ref{#1}}

\def\1{\bm{1}}

\def\va{{\bm{a}}}

\def\vh{{\bm{h}}}

\def\vx{{\bm{x}}}
\def\vy{{\bm{y}}}
\def\vz{{\bm{z}}}

\DeclareMathAlphabet{\mathsfit}{\encodingdefault}{\sfdefault}{m}{sl}
\SetMathAlphabet{\mathsfit}{bold}{\encodingdefault}{\sfdefault}{bx}{n}

\def\gT{{\mathcal{T}}}

\usepackage{xspace}
\usepackage{tabularx}
\usepackage{booktabs}
\newcolumntype{Y}{>{\centering\arraybackslash}X}
\newcolumntype{L}{>{\raggedright\arraybackslash}X}
\def\gT{{\mathcal{T}}}
\newcommand{\ttbf}[1]{\texttt{\textbf{#1}}}
\newcommand{\da}[0]
{\ttbf{PlanSDG}\xspace}

\title{\emph{Turning Flowchart into Dialog}: Augmenting Flowchart-grounded Troubleshooting Dialogs via Synthetic Data Generation}

\author{Haolan Zhan, \quad Sameen Maruf, \quad {Lizhen Qu}, \quad Yufei Wang \\
\textbf{Ingrid Zukerman} \and \textbf{Gholamreza Haffari} \\
 Department of Data Science \& AI, Monash University, Australia \\
\{firstname.lastname\}@monash.edu\\ 
}

\begin{document}
\maketitle
\begin{abstract}

Flowchart-grounded troubleshooting dialogue (FTD) systems, which follow the instructions of a flowchart to diagnose users' problems in specific domains (\eg\ vehicle, laptop), have been gaining research interest in recent years. However, collecting sufficient dialogues that are naturally grounded on flowcharts is costly, thus FTD systems are impeded by scarce training data. To mitigate the data sparsity issue, 
we propose a plan-based  
synthetic data generation (\da) approach
that generates diverse synthetic dialog data at scale by transforming concise flowchart into dialogues.
Specifically, its generative model employs a variational-base framework with a hierarchical planning strategy that includes \textit{global} and \textit{local} latent planning variables. Experiments on the {FloDial} dataset show that synthetic dialogue produced by \da improves the performance of downstream tasks, including flowchart path retrieval and response generation, in particular on the \textit{Out-of-Flowchart} settings. In addition, further analysis demonstrate the quality of synthetic data generated by \da in {paths that are covered by current sample dialogues and paths that are not covered.}

\end{abstract}

\section{Introduction}
\label{intro}
\begin{figure}[!t]
    \centering
    \includegraphics[width=0.99\linewidth]{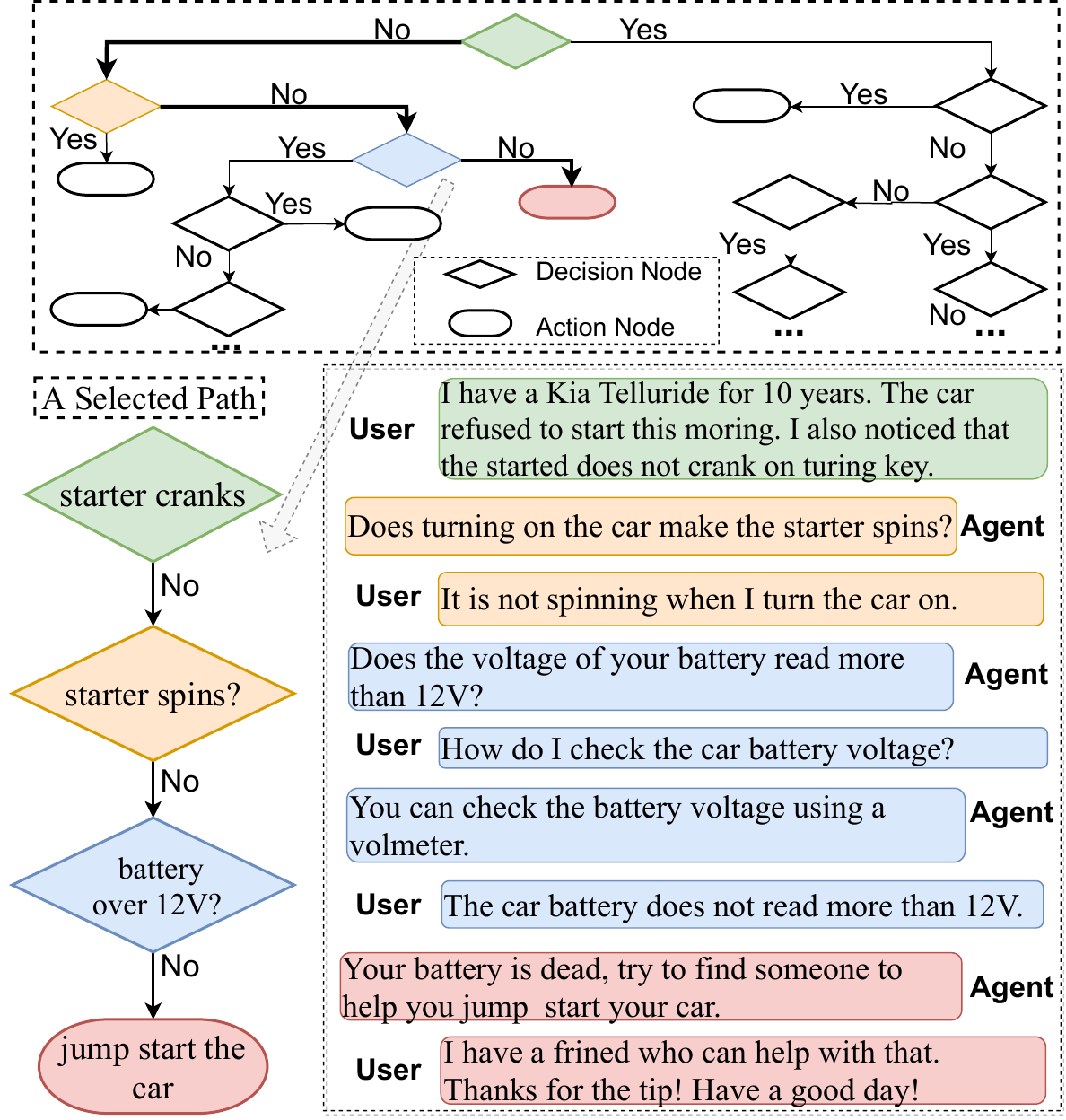}
    \caption{A sample flowchart-grounded troubleshooting dialogue. Agent follows the path of a flowchart to help user diagnose  problems.}
    \label{fig:intro}
\end{figure}

{\em Flowchart-grounded Troubleshooting Dialogue } ({\em FTD}\/) systems~\cite{leake2005using,boye2007dialogue,williams2007applying,paek2008automating,janarthanam2008user,wei2018task,flowchart2021emnlp}, which communicate with users to help them diagnose problems through the guidance of a flowchart, have been gaining interest in recent years.
FTD systems face additional challenges to those faced by typical task-oriented dialogue systems~\cite{wen2017network,budzianowski2018multiwoz},
\eg\ 
FTD systems must accurately follow the instructions of a flowchart, actively detect the 
root cause of issues, and provide users with reasonable solutions by following an action instruction along with the \textit{path} in a flowchart (Figure~\ref{fig:intro}).

%

\begin{figure}[!t]
    \centering
    \includegraphics[width=1.0\linewidth]{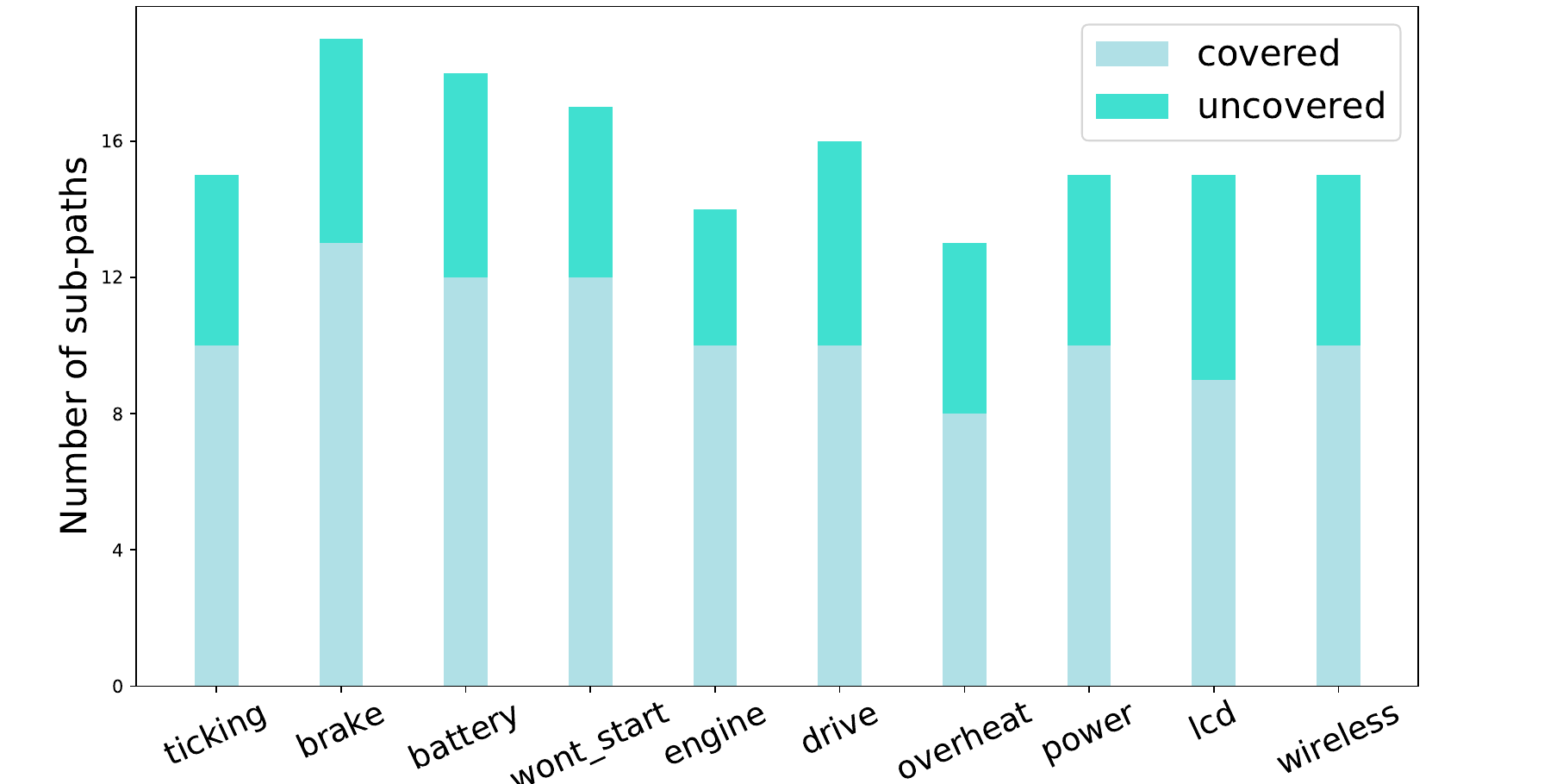}
    \caption{Statistics on the percentage (\%) of (un)covered paths in the \textit{FloDial} (containing ten flowcharts in two domains: Vehicle and Laptop) -- each flowchart pertains to a specific problem. In total, more than 35\% of paths are not covered by dialogue instances.}
    \label{fig:stat}
    \vspace*{-3mm}
\end{figure}

Collecting sufficiently large flowchart-related dialogue corpora for FTD is challenging, 
since it requires domain experts with relevant knowledge. This problem also applies to a
crowd-sourced FTD corpus, such as \Flodial~\cite{flowchart2021emnlp}, whose
collection still involved a great deal of human effort. 
Despite this, the 1,789 dialogues in \Flodial\ (\S~\ref{section:expsetup}) cover only 65\% of the paths in the
underlying flowcharts on average (Figure~\ref{fig:stat}).
%
An alternative approach to obtaining additional dialogues could involve crawling through websites. 
However, most of these data obtained in this manner focus on anecdotes and subjective opinions~\cite{dai2022dialoginpainting}, and are thus unsuitable for FTD systems.


In this paper, we propose \da: a \textbf{Plan}-based 
\textbf{S}ynthetic \textbf{D}ata \textbf{G}eneration 
approach that generates synthetic dialogues from flowchart paths.
Specifically, \da takes as input a path extracted from an underlying flowchart, and generates a dialogue session consisting of dialogue acts and utterances. 
\da is formalised as a probabilistic generative model with structured planning latent variables, specifically
\textit{global} and \textit{local} latent variables, that guide the generation process.
The \textit{global} latent variables are responsible for modeling the dialogue acts between  the dialogue turns, 
providing a high-level sketch. 
To be able to model these global variables, we manually labeled the dialogue acts for the utterances in the \Flodial\ dataset.
The \textit{local} latent variables control the diversity of generated synthetic dialogues during sentence realization.

We conducted \textit{extrinsic} and \textit{intrinsic} evaluations of our approach  on the \Flodial\ corpus, as well as 
follow-up ablation studies.
Our extrinsic evaluation shows that the retrieval and generative models trained on the synthetic 
dialogues produced by \da achieve better performance than other augmentation methods in terms of the downstream tasks: flowchart path retrieval and response generation, 
particularly on the \textit{Out-of-Flowchart} settings.
Our intrinsic evaluation, which examines the quality of the synthetic dialogues, 
indicates that \da outperforms strong baseline models in term of diversity and faithfulness.
Our ablation studies demonstrate the effectiveness of our proposed \textit{global} 
and \textit{local} latent planning variables. 
Further analysis demonstrate the quality of synthetic data generated by \da in \textit{uncovered paths} that are included by flowchart but not in  dialogues.



\section{Plan-based Synthetic Data Generation}

\subsection{Task Formulation}
\label{formulation}
The goal of \da is to take a sampled path from the flowchart, and generate a complete synthetic dialogue as well as the dialogue acts. In this paper, we only have access to a (relatively small) training set  
$\gT=\{(\vx,\va,\vy)_i\}_{i=1}^m$, where ${\vx} = \{{x}_1, {x}_2, \ldots, {x}_n\}$ is a flowchart {path}. 
A path includes tuples of nodes and edges from the flowchart.
Each  $x_i \in \vx$ on the path corresponds to a sub-dialogue $y_i=[y_{i, 0}, \cdots, y_{i, |y_i|}] \in \vy$, where $y_{i,j}$ is an utterance associated with a dialogue act $a_{i,j} \in \va$. 
For example in the flowchart path in Figure \ref{fig:intro}, the node ``battery over 12V'' ($x_3$) corresponds to the sub-dialogue starting from the turn ``Does the voltage of $\ldots$'' and ending to the turn ``The car battery does not $\ldots$'' ($y_{3,0}$ to $y_{3,3}$), where each turn is associated with a dialogue act.

Given a flowchart {path} $\vx$, our proposed {data augmentation} method \da generates synthetic dialogue acts $\hat{\va}$ and dialogues turns $\hat{\vy}$, and produces the synthetic dataset $\gT_{Syn}=\{(\vx, \hat{\va},\hat{\vy})_i\}_{i=1}^n$ where $n$ could be much larger than $m$ (\eg\ 10x). Our goal is that the downstream retrieval and generative dialogue models trained using $\gT \cup \gT_{Syn}$ outperform the models trained using only $\gT$.



\begin{figure*}
\centering
\includegraphics[width=1\linewidth]{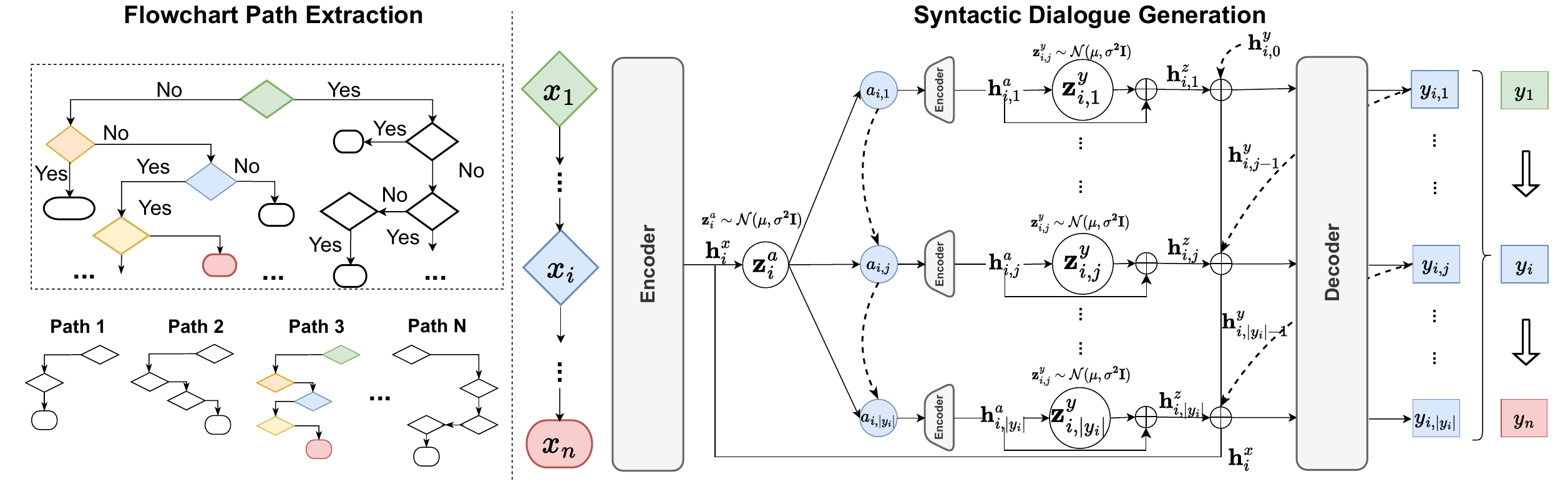}
\label{screenshot}
\vspace{-3mm}
\caption{Detailed framework of \da, including  path extraction and synthetic dialogue generation.}
\label{fig:framework}
\end{figure*}

\subsection{Flowchart Path Extraction}
\label{pathextract}
As shown in Figure~\ref{fig:intro}, the flowcharts used in this paper consist of decision nodes and action nodes. The decision nodes include a question, and they are connected with other nodes by the user responses (\eg\ Yes, No). 
The action nodes at the bottom of the flowcharts indicate the recommended actions. 

{For training \da, we directly extract the flowchart paths for the dialogues in the training set.} 
{For syntactic data generation,}
to ensure full coverage for the flowchart paths, we extract the flowchart paths by \emph{Depth-First-Search} from the top decision node to the bottom action nodes. The resulting flowchart paths are then used as the inputs for \da.

\subsection{Synthetic Dialogue Generation}
\label{overviewda}
\da is designed to generate diverse and high-quality synthetic dialogues from the extracted flowchart paths. Even though the input flowchart paths include textual questions, user responses and final actions, conditioning only on this information could result in tedious conversations 
{consisting of rigid sequences of question-answer pairs.} 
Starting from a node in a flowchart, there could be many feasible open-ended dialogues. 
{To facilitate coverage of this dialogue space, we employ intermediate latent variables in \da. Dialogue
acts are an intuitive choice to characterise these variables, as they
describe the basic function of a dialogue turn/utterance (\eg\ inform, clarification), and reflect users' intentions~\cite{stolcke2000dialogue,bunt2011semantics}.
We denote them by global latent variables $\vz_i^a$, responsible for modeling the dialogue act transition process \emph{over the turns}. 
We further introduce local latent variables $\vz_{ij}^y$, responsible for generating lexically diverse utterances for each turn. 
As such, \da is formally a probabilistic generative model with structured latent variables (Figure~\ref{fig:framework}), explained below in more details. }


\paragraph{Global Planning over Dialogue Acts.}
We inject stochasticity into the global planning process using a continuous latent variable in each dialogue turn
$\vz^a_i$, which is assumed to follow the isotropic Gaussian distribution~\cite{kingma2014auto}. We first sample $\mathbf{z}^a_i$ from its prior distribution $p_\theta^{\vz^a} (\vz^a_i | x_i)$, and then generate a sequence of dialogue acts auto-repressively:
\begin{align}
    \vz^a_i&~\sim~p_\theta^{\vz^a} (\vz^a_i | x_i) \\
    a_{i,j}&=p^{a}_{\theta}(. | a_{i,j-1}, x_i,\vz^a_i)
\end{align}
where $p_{\theta}^a(a_{i,j} |a_{i,j-1}, {\vz}^{ a}_i,\vh^{x}_i)$ is a 2-layer MLP with the softmax on top. We train $p_\theta^{\vz^a} (\vz^a_i | x_i)$ to approximate the  posterior distribution $q_\phi (\vz^a_i  | x_i, y_i)$ using Gaussians in the training phase. The parameters in the prior and posterior distributions, $\bm{\mu}^{p}_{a}$, $\bm{\sigma}^{p}_{a}$,  $\bm{\mu}^{q}_{a}$ and $\bm{\sigma}^{q}_{a}$, are parameterised as follows:
\begin{align*}
    \bm{\mu}^p_{a} &= \mathit{MLP}^p_\theta({\vh^{x}_i}), \\
    \bm{\sigma}^p_{a} &= \mathit{softplus}(\mathit{MLP}^p_\theta(\vh^{x}_i)), \\
    \bm{\mu}^q_{a} &= \mathit{MLP}^q_\phi([{\vh^{x}_i,\vh^{y}_i]}), \\
    \bm{\sigma}^q_{a} &= \mathit{softplus}(\mathit{MLP}^q_\phi([\vh^{x}_i,\vh^{y}_i])), 
\end{align*}
where ${\rm MLP}(\cdot)$ denotes a multi-layer perceptron, ${\rm softplus(\cdot)}$ is a smooth approximation to ReLU, which ensures positiveness. $\vh^x_i = \mathit{AvgPool}(\mathit{Enc}(x_i))$ and $\vh^{y}_i=\mathit{AvgPool}(\mathit{Enc}([y_{i, 0}, \cdots, y_{i, k}]))$, which allows $\vz^a_i$ to capture the global utterance information associated with $x_i$. Finally, the Evidence Lower Bound (ELBO) is computed as follows:
\begin{eqnarray*}
  \lefteqn{\mathcal{L}_{\mathrm{global}}  =  -D_{KL}(q_\phi({\vz}^{a}_i|{x}_i, {y}_i)||p_\theta^{\vz^a}({\vz}^{a}_i|{x}_i))} \\
    & & + \mathbb{E}_{{\vz}^{a}_i \sim q_\phi}[\sum_j \log p_{\theta}^a({a_{i,j}} |a_{i,j-1},{\vz}^{ a}_i,{x}_i)],
\end{eqnarray*}
where $D_{KL}(\cdot |\cdot)$ denotes the Kullback-Leibler divergence~\cite{kullback1951information}.

\paragraph{Local Planning for Utterance Generation.}
Given the dialogue act $a_{i,j}$ generated from $\vz^a_i$, we focus on generating lexically diverse dialogue utterances that are faithful to the flowchart. We sample $\mathbf{z}^y_{i,j}$ from its prior distribution conditioned on $a_{i,j}$ and $x_i$, as follows:
\begin{align}
    \vz^{y}_{i,j}~\sim~p^{\vz^y}_{\theta}(\vz^{y}_{i,j}| x_i, a_{i,j})
\end{align}
We train $p^{\vz^y}_{\theta}(\vz^{y}_{i,j}| x_i, a_{i,j})$  to approximate the  posterior distribution $q_\phi (\vz^y_{i,j} | x_i, a_{i,j}, y_{i,j})$, assuming that both distributions are Gaussian. They are parameterised as follows:
\begin{align*}
\bm{\mu}^p_{y} &= \text{MLP}^p_\theta({\vh}^{x}_i,\vh^a_{i,j}), \\
\bm{\sigma}^p_{y} &= \text{softplus}(\text{MLP}^p_\theta({\vh}^{x}_i,\vh^a_{i,j})\\
\bm{\mu}^q_{y} &= \text{MLP}^q_\phi({\vh}^{x}_i,\vh^a_{i,j},\vh^y_{i,j}), \\ \bm{\sigma}^q_{y}
&= \text{softplus}(\text{MLP}^q_\phi({\vh}^{x}_i,\vh^a_{i,j},\vh^y_{i,j})),
\end{align*}
where $h^a_{i,j}=\mathit{AvgPool}(\mathit{Enc}(a_{i,j}))$. In contrast with global planning, here we use the ground-truth utterance $y_{i,j}$ for training to allow \da to focus on the local information. 
Finally, the ELBO for the local planning variable is:
\vspace*{-2mm}
\begin{eqnarray*}
    \lefteqn{\mathcal{L}_{\mathrm{local}} =}\\
    & & \!\!\!-D_{KL}(q_\phi({\vz}^{y}_{i,j}|{x}_i,a_{i,j}, y_{i,j})||p_\theta^{\vz^y}({\vz}^{y}_{i,j}|{x}_i,a_{i,j})) \\
    & & \!\!\!+\mathbb{E}_{{\vz}^{y}_{i,j} \sim q_\phi}[\log p_{\theta}({y}_{i,j} |y_{i,j-1}, x_i, a_{i,j}, \vz^y_{i,j})].
\end{eqnarray*}
%
%
\da generates each utterance $y_{i,j}$ based on $\vh^z_{i,j}$, $x_i$ and $y_{i,j-1}$, as follows:
\begin{align*}
    y_{i,j}  = \mathit{Dec}(\vh^y_{i,j-1}, \vh^x_i, \vh^z_{i,j}),
\end{align*}
where $\vh^z_{i,k}=\text{Concat}([\vh^a_{i,j}, \vz^{y}_{i,j}])$ is the concatenation of the global and local planning variables. 
$\mathit{Enc}$ and $\mathit{Dec}$ are based on the Transformer architecture, and their parameters are initialized from a pre-trained Seq2Seq model (\eg\ BART). 

\subsection{Training Objective}
\label{trainingobj}
To summarise, the probabilistic generative model of \da performs the following steps to produce a dialogue from a flowchart path $\vx$. 
For each $x_i \in \vx$ on the path, it starts by sampling the global latent variable $\vz^{a}_i \sim p^{\vz^a}_{\theta}(. | x_i)$, and then iteratively samples the turns $y_{i,j}$ as follows:
\begin{itemize}
    \myitem Sample the  dialogue act:\\  $a_{i,j}~\sim~p^{a}_{\theta}(. | a_{i,j-1}, x_i,\vz^a_i)$
    \myitem Sample the local latent variable:\\
$\vz^{y}_{i,j}~\sim~p^{\vz^y}_{\theta}(. | x_i, a_{i,j})$   
\myitem Sample the utterance:\\
 $y_{i,j}~\sim~p^y_{\theta}(.|y_{i,j-1}, x_i, a_{i,j}, \vz_{i,j}^y)$
\end{itemize}
Hence, the probability of generating a conversation and the corresponding dialogue acts given the flowchart path can be written as follows:
\begin{eqnarray}
\label{eqn:probgen}
\lefteqn{p_{\theta}(\vy,\va|\vx) = \prod_i  \int d(\vz_i^a)  p_{\theta}^{\vz^a}(\vz_i^a|x_i)}  \\  
    &&  \times \prod_j \int d(\vz_{i,j}^y)    p_{\theta}^a(a_{i,j}|a_{i,j-1},x_i, \vz_i^a) \nonumber \\ 
    && \times p_{\theta}^{\vz^y}(\vz_{i,j}^y|x_i,a_{i,j}) p^y_{\theta}(y_{i,j}|y_{i,j-1}, x_i, a_{i,j}, \vz_{i,j}^y) \nonumber
\end{eqnarray}
The overall training objective of \da is the  sum of the ELBOs:
$\mathcal{L} = \mathcal{L}_{\mathrm{global}} + \mathcal{L}_{\mathrm{local}}.$
This is based on the variational approach  to overcome the
challenges of integration over the latent variables in the likelihood objective  (Equation~\ref{eqn:probgen}).
We use the re-parametrization
trick in~\cite{kingma2014auto} to optimise the training objective.

\section{Experiments}

\subsection{Setup}
\label{section:expsetup}
\paragraph{Dataset} 
We use the \textit{FloDial} dataset~\cite{flowchart2021emnlp} for our experiments. \textit{FloDial} is a  troubleshooting dialogue corpus containing 1,789 dialogues grounded on ten individual flowcharts\footnote{There is no path interaction or overlap between two individual flowcharts.}  from two main domains: vehicle and laptop (five flowcharts in each domain).
\textit{FloDial} has two different settings: \textit{In-Flowchart} and \textit{Out-of-Flowchart}.
In the \textit{In-Flowchart} setting, both the training and test data are grounded on the same sets of flowcharts, while in the \textit{Out-of-Flowchart} setting, the test dialogues are based on the flowcharts that are not included in the training stage.



        

\begin{figure}
    \centering
    \includegraphics[scale=0.35]{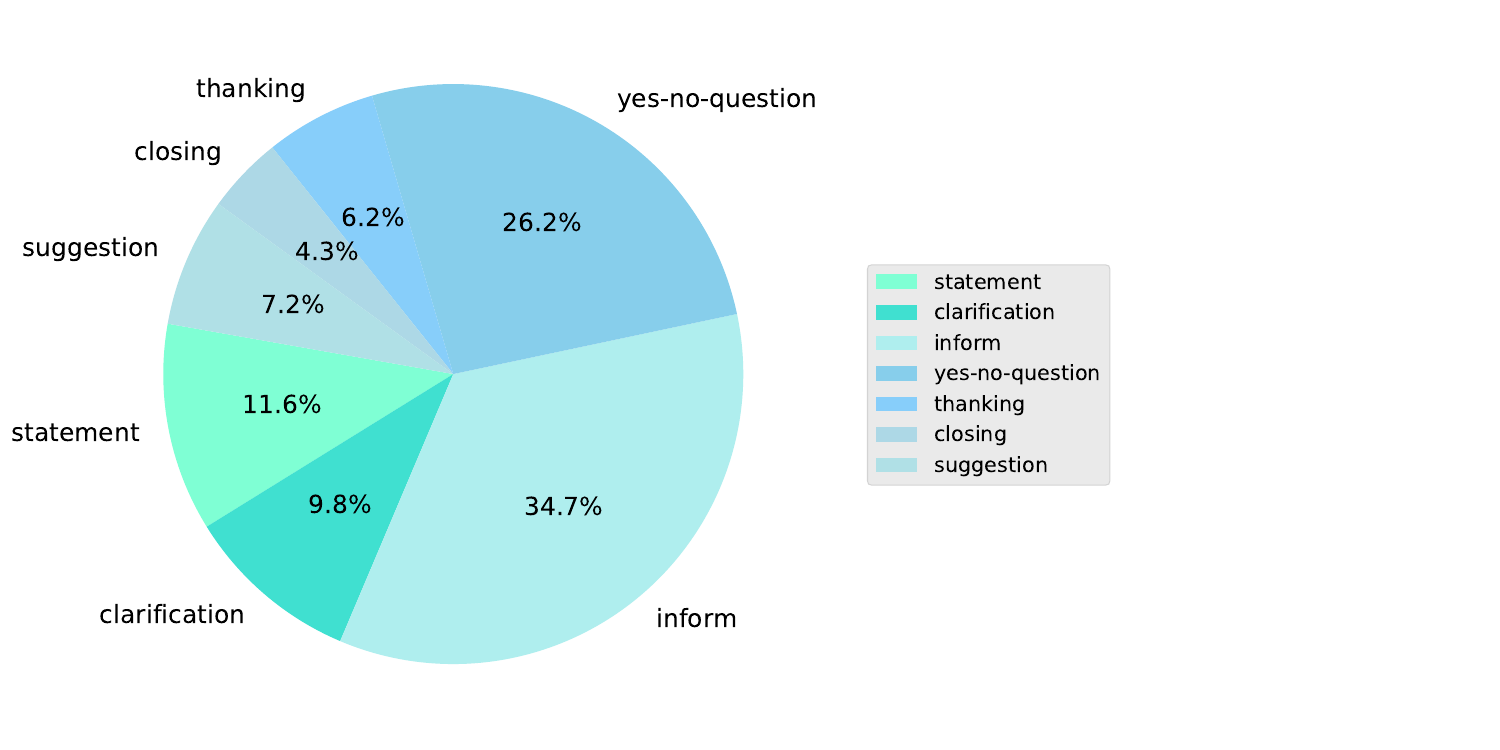}
    \caption{Statistics of dialogue act proportions in the \Flodial\ dataset.}
    \label{fig:dialogAct}
\end{figure}

\paragraph{Dialogue Act Labeling}
As the original \textit{FloDial} dataset does not contain dialogue act labels, we manually label the dialogue act for each utterance.
We investigated several widely-used dialogue act datasets, including Switchboard\footnote{\url{https://catalog.ldc.upenn.edu/LDC97S62}}, AMI\footnote{\url{https://groups.inf.ed.ac.uk/ami/corpus/}} and MultiWoz.\footnote{\url{https://github.com/budzianowski/multiwoz}}
From these datasets, we select the most commonly used set of dialogue acts (i.e., cover 74.38\% of the dialogue acts in these datasets) that are compatible with the \textit{FloDial} dataset, including \{\textit{statement}, \textit{inform}, \textit{yes-no-question}, \textit{clarification}, \textit{thanking}, \textit{closing}, \textit{suggestion}\}, and conduct annotation\footnote{\url{https://github.com/zhanhl316/flowchart-dialogue-with-DA}} for the \textit{FloDial} dataset.
Figure~\ref{fig:dialogAct} shows the detailed statistics of the labeled dialogue acts.

\begin{table*}[t]
\small
\centering
\begin{tabularx}{\textwidth}{lYYYYcYYYY}
\bottomrule
\multirow{2}{*}{Augmentation} & \multicolumn{4}{c}{\textit{In-Flowchart} } && \multicolumn{4}{c}{\textit{Out-of-Flowchart} }  \\ 
 \cmidrule{2-5} \cmidrule{7-10}
{Model} & PPL $\downarrow$ & \small{BLEU} $\uparrow$ & R@1 $\uparrow$ & R@5 $\uparrow$  && PPL $\downarrow$ & \small{BLEU} $\uparrow$ & R@1 $\uparrow$ & R@5 $\uparrow$ \\ 
\midrule
\rowcolor{black!15} FloNet & 4.93 & 19.36 & 0.834 & 0.957 && 17.08 & 9.53 & 0.529 & 0.765 \\
\midrule
EDA &  5.67 & 19.65 &  0.837 & 0.956 && 16.84 & 9.79 & 0.535 & 0.772 \\
Back-Tran &  4.88 & 19.93 & 0.839 & 0.952 && 19.26 & 10.67 &  0.538 & 0.781 \\
GPT-2 & 4.37 & 20.69 & 0.844 & 0.958 && 15.93 & 13.70 & 0.574 & 0.813 \\ 
BART & 4.52 & 21.11 & 0.852 & 0.965 && {12.48} & 13.94 & 0.581 & 0.826\\ 
\midrule
\da w/o $\mathcal{L}_{\mathrm{global}}$ & 4.61 & 20.75 & 0.847 & 0.963  && 14.25 & 14.17 & 0.583 & 0.829 \\
\da w/o $\mathcal{L}_{\mathrm{local}}$ & 4.48 & 21.06 & 0.843 & 0.956  &&  \textbf{12.45} & 13.83 &  0.579 & 0.832\\
\da & \textbf{4.35}$^{*}$ & \textbf{21.18}$^{*}$ & \textbf{0.853}$^{*}$ & \textbf{0.968}$^{*}$ && {12.64} & \textbf{14.73}$^{**}$ & \textbf{0.609}$^{**}$ & \textbf{0.841}$^{**}$ \\
\midrule
DialoGPT & 4.19 & 20.93 & 0.849 & 0.961 && 14.66 & 12.63 & 0.557 & 0.793 \\ 
BlenderBot & \textbf{4.06} & \textbf{21.26}& 0.847 & 0.960  && 13.06 & 12.89 & 0.562 & 0.804 \\ 
\bottomrule
\end{tabularx}
\caption{Extrinsic evaluation: Performance of augmented synthetic dialogue data generated by different models in \textit{In-Domain} and \textit{Out-of-Domain} settings. Results are based on the augmentation of \textbf{10x} the amount of data. Scores marked with ``$^\star$'' and ``$^{\star\star}$'' respectively indicate a significance of $\pvalue < 0.05$ and $\pvalue < 0.01$ in the t-test after Benjamini-Hochberg (BH) correction for false discovery rate~\cite{benjamini1995controlling}.}
\label{tab:aug_baseline} 
\end{table*}

\paragraph{Evaluation Settings}
In this paper, we conduct following evaluation: \textbf{1) Extrinsic Evaluation}: We aim to verify whether the synthetic data generated from the baselines and \da are useful for improving the performance of FTD. To precisely measure FTD performance, we use the same evaluation metrics as \citet{flowchart2021emnlp}: Perplexity (PPL) and BLEU~\cite{papineni2002bleu} for response generation, and R@1 and R@5 for flowchart node retrieval.\footnote{In order to diagnose problems, at each step, the agent must retrieve the most relevant node from flowchart database.} \textbf{2) Intrinsic Evaluation}: We aim to confirm if our proposed model \da generate more diverse and faithful pseudo-dialogues than the baseline models. To investigate the quality of generated synthetic data from \da and other baseline models, we use ROUGE~\cite{lin2004rouge} to assess fluency, Distinct~\cite{li2016distinct} and {Self-BLEU}~\cite{zhu2018selfbleu} for diversity, and Embedding Metrics (Average, Extrema, Greedy) and BART-Score~\cite{yuan2021bartscore} for faithfulness.

\paragraph{Baselines} 
Our baseline is \textbf{FloNet}~\cite{flowchart2021emnlp} 
which only uses the original training data $\gT$. Given the newly generated synthetic data $\gT_{Syn}$ from \da and other synthetic data generation models, we train the same \textbf{FloNet} model with $\gT \cup \gT_{Syn}$ under the same set of hyper-parameters. We compare \da with the following synthetic data generation models:
\begin{itemize}
    \item \textbf{EDA}~\cite{wei-zou-2019-eda} is a rule-based approach by synonym replacement, random insertion, random swap, and random deletion.
    \item \textbf{Back-Tran}~\cite{sennrich2016improving} is the classical back translation algorithm rooted from the machine translation task.
    \item Generic pre-trained language models including \textbf{GPT-2}~\cite{radford2019language}, \textbf{BART}~\cite{lewis2019bart}.
    \item Conversational pre-trained models including \textbf{DialoGPT}~\cite{zhang2019dialogpt} and \textbf{BlenderBot}~\cite{roller2020recipes}.
\end{itemize}
We use the large version for all pre-trained models. To make a fair comparison, we incorporate annotated dialogue acts for both \da and other synthetic data pre-trained models.



\paragraph{Implementation Details}
We utilize the state-of-the-art pre-trained text generation model BART to initialize the encoder and decoder of \da, for both prior and posterior, encoder and generator. For fair comparison with baseline models, we use the ${\rm BART}_{\rm large}$ for our model. In preliminary experiments, we find that fine-tuning outperforms prompt-tuning~\cite{li-liang-2021-prefix} for generating valid dialogue data. For training process, we use AdamW~\cite{loshchilov2017decoupled} for gradient optimization, learning rate 0.001. batch size 8 in our experiments. We fine-tune \da for 50 epochs and the maximum length for utterances is set to 64. To mitigate the posterior collapse issue, we adopt the KL thresholding strategy~\cite{kingma2016improved} that maximizes the KL term with a constant $\beta = 0.1$\footnote{The code will be made available upon publications.}.

\begin{table*}[!ht]
\small
\centering
\begin{tabularx}{\textwidth}{cYYYYcYYYY}
\bottomrule
\multirow{2}{*}{Data Size} & \multicolumn{4}{c}{\textit{In-Flowchart} } && \multicolumn{4}{c}{\textit{Out-of-Flowchart} }  \\ 
 \cmidrule{2-5} \cmidrule{7-10}
 & PPL $\downarrow$ & \small{BLEU} $\uparrow$ & R@1 $\uparrow$ & R@5 $\uparrow$  && PPL $\downarrow$ & \small{BLEU} $\uparrow$ & R@1 $\uparrow$ & R@5 $\uparrow$ \\ 
\midrule
\rowcolor{black!15} FloNet (1x) & 4.93 & 19.36 & 0.834 & 0.957 && 17.08 & 9.53 & 0.529 & 0.765  \\ 
       2x Data & 5.26 & 20.72 & 0.843 & 0.956 && 13.27$^{**}$ & 11.75$^{*}$ & 0.546$^{**}$ & 0.819$^{**}$ \\ 
       5x Data & 4.28 & 21.06$^{*}$ & 0.851$^{*}$ & 0.961 && 15.63$^{**}$ & 14.01$^{**}$ & 0.595$^{**}$ & 0.837$^{**}$ \\ 
       10x Data & 4.35$^{*}$ & 21.18$^{*}$ & 0.853$^{*}$ & 0.968 && 12.64$^{**}$ & 14.73$^{**}$ & 0.609$^{**}$ & 0.841$^{**}$\\ 
\bottomrule
\end{tabularx}
\caption{Extrinsic performance. FloNet (1x) is the dataset of the baseline model~\cite{flowchart2021emnlp}. 2x, 5x and 10x means that we extend the original \textit{FloDial} training set with different amounts of synthetic data. Scores marked with ``$^\star$'' and ``$^{\star\star}$'' indicate a significance of p < 0.05 and p < 0.01 in the t-test with BH correction respectively.}
\label{tab:aug_data_size}
\end{table*}


\subsection{Extrinsic Evaluation}


\paragraph{Main Results} 
Table~\ref{tab:aug_baseline} summarizes the augmentation experiment results using 10 times (10x) for both baseline data augmentation models and \da. In both settings, the performance of response generation and flowchart node retrieval tasks trained with the synthetic data from \da are boosted up, especially in the \textit{Out-of-Flowchart} setting. Specifically, \da outperforms rule-based \textbf{EDA} and naive \textbf{Back-Tran} methods by a large margin, demonstrating that widely-used data augmentation methods cannot handle the FTD situations.  While comparing with strong pre-trained models (e.g, GPT-2, BART), synthetic data generated by our model have better augmentation performance. We see that \da is more effective in the \textit{Out-of-Flowchart} setting, 
though it is on-par or better than the baselines in the  \textit{In-Flowchart} setting. In the \textit{out-of-Flowchart} setting, \da achieves at least 5.6\% and 4.8\% for BLEU and R@1 metric than baseline models. Surprisingly, model performance supported by \da even surpass those models supported by DialoGPT and BlenderBot which use large-scaled dialogue data for pre-training. This result suggests that with small training data, \da can generalize well to the domains not encountered (i.e., dialogue) in its pre-training stage.

\paragraph{Analysis on Synthetic Data Size} 
Table~\ref{tab:aug_data_size} presents the augmentation performance using different size of synthetic data. FloNet (1x) only uses original training data. As shown in Table~\ref{tab:aug_data_size}, the FloNet model performance keeps improving along with the data size expansion. Especially in the \textit{Out-of-Flowchart} setting, augmentation performance improve significantly comparing to the FloNet (1x) model. These results demonstrate that \da can effectively learn from existing training data and produce diverse and relevant synthetic data rather than introducing noise information. 

\begin{table}[t]
\small
\centering
\begin{tabularx}{0.5\textwidth}{lYYYY}
\bottomrule
\multirow{2}{*}{Model} & \multicolumn{4}{c}{Uncovered path within flowchart }  \\ 
 \cmidrule{2-5} 
 & PPL $\downarrow$ & \small{BLEU} $\uparrow$ & R@1 $\uparrow$ & R@5 $\uparrow$  \\ 
\midrule
\rowcolor{black!15} FloNet & 12.94 & 11.05 & 0.597 & 0.815 \\
\midrule
EDA & 12.36 & 11.69 & 0.598 & 0.804 \\
Back-Tran & 13.67 & 12.18 & 0.608 & 0.827 \\
GPT-2 & 9.82 & 14.61 & 0.632  & 0.854 \\ 
BART & 8.46 & 15.29 & 0.637 & 0.852 \\ 
\midrule
\da & \textbf{8.26}$^{*}$ & \textbf{15.90}$^{**}$ & \textbf{0.654}$^{**}$ & \textbf{0.868}$^{*}$ \\
\bottomrule
\end{tabularx}
\caption{Augmentation performance on \textit{Uncovered path} in the flowchart (\textit{In-Flowchart} using 10x augmented synthetic data.). Scores marked with ``$^\star$'' and ``$^{\star\star}$'' indicate a significance of p < 0.05 and p < 0.01 in the t-test with BH correction respectively.}
\label{tab:aug_uncovered} 
\end{table}

\paragraph{Analysis on Uncovered Path} 
To verify the effectiveness of \da on uncovered path, we conduct additional experiments on a novel uncovered path setting. As discussed above, the existing training data only cover 65\% of the flowchart path in the \textit{FloDial} dataset. We split these training datasets into training (80\%), as covered path, and testing (20\%), as uncovered path. Table~\ref{tab:aug_uncovered} summarizes the results on the uncovered path setting. \da achieves the best augmentation performance comparing to other augmentation baseline models. The positive results demonstrate that \da is capable enhance the model performance on those uncovered flowchart paths.


\paragraph{Ablation on Latent Variables} 
We conduct ablation study for the components of \textit{local} and \textit{global} planning variables described in Section~\ref{overviewda}. As shown in Table~\ref{tab:aug_baseline}, the elimination of \textit{local} and \textit{global} planning variables  undermine the performance of \da, showing the positive contribution of these two latent variables in generating diversity and relevant synthetic data. Specifically, the ablation of \textit{local} planning variable leads to more performance degradation than the ablation of \textit{global} in terms of flowchart node retrieval task, showing the importance of \textit{local} variable in controlling the diversity on sentence realization, which further impact the training on downstream tasks.

\begin{table*}[t]
\small
\centering
\begin{tabularx}{\textwidth}{lYYYYYYc}
\bottomrule
Model   & \small{BU-4} $\uparrow$ & \small{RG-L} $\uparrow$  &  \small{Dist-2} $\uparrow$ & \small{Dist-3} $\uparrow$  & \small{Self-B} $\downarrow$  & \small{BART-S} $\downarrow$ & \small{Emb (Avg/Extr/Gre)} $\uparrow$  \\
\midrule
GPT-2   & {26.8} & {43.1} & 0.267 & 0.425 & 0.328 & -2.590 &  88.1/68.7/84.1  \\
  BART   & \textbf{29.7} & {47.2}  & 0.351 & 0.541 & 0.271 & -2.164 & 87.2/67.5/83.3  \\
  DialoGPT  & {24.7} & {40.1}  & 0.366 & 0.563 & 0.257 & -2.328 & \textbf{89.3}/62.5/82.6  \\
  BlenderBot  & {19.3} & {35.6}  & 0.308 & 0.497 & 0.283 &  -2.051 &  82.6/59.3/78.6  \\
  \midrule
  w/o $\mathcal{L}_{\mathrm{global}}$  & 27.3 & 49.1 & 0.382 & 0.574 & 0.249 & -2.156 &{87.1}/68.3/84.7 \\
  w/o $\mathcal{L}_{\mathrm{local}}$  & 27.8 & 47.6 & 0.365 & 0.568 & 0.261 & -2.321 & 85.7/68.2/83.8  \\
  \da  & {28.5} & \textbf{51.2}$^{\star\star}$  & \textbf{0.397}$^{**}$ & \textbf{0.602}$^{**}$ & \textbf{0.225}$^{**}$ & \textbf{-2.037}$^\star$ &  86.1/\textbf{69.4}$^\star$/\textbf{85.7}$^{**}$ \\
\bottomrule
\end{tabularx}
\caption{Intrinsic evaluation results for pseudo dialogue generation. The metrics  BLEU-4, ROUGE-L, Distinct-2/3, Self-BLEU, BART-score and Embedding are
abbreviated as BU-4, RG-L, Dist-2/3, Self-B, BART-S and Emb respectively. The best results are highlighted with \textbf{bold}. Scores marked with ``$^\star$'' and ``$^{\star\star}$'' indicate a significance of p < 0.05 and p < 0.01 in the t-test with BH correction respectively.}
\label{tab_auto-eval}
\end{table*}


\subsection{Intrinsic Evaluation}
In this section, we directly verify the quality of synthetic data by using various of automatic metrics.

\paragraph{Automatic Metrics}
We show the automatic intrinsic evaluation results on synthetic dialogue in Table~\ref{tab_auto-eval}. 
\da outperforms the baselines in terms of ROUGE-L, Dist-2/3, Embedding and BART-Score. For BLEU-4 the results of \da are close to the baseline models. The significant improvement obtained by \da for Dist-2/3 indicates that our model is able to generate more diverse texts than the baselines -- a result of our latent variable modeling. The high scores of Embedding and BART-Score indicate that our model also has the capacity to generate utterances that are semantically coherent with the input flowchart.

\paragraph{Ablation on Latent Variables}
We first show the ablation study of different training objectives in Table~\ref{tab_auto-eval}. We observe a certain performance drop when removing global planning latent variable $\mathcal{L}_{\mathrm{global}}$ or local planning latent variable $\mathcal{L}_{\mathrm{local}}$  during fine-tuning. Specifically, the removal of $\mathcal{L}_{\mathrm{local}}$ results in a significant drop in Dist-2/3 metric, showing that the local planning latent variable, together with dialogue act, is responsible for utterance diversity. We then highlight that the significance of dialogue act plays an important role in high-level sketch. The absence of $\mathcal{L}_{\mathrm{global}}$ also results in a drop of performance in terms of BLEU-4, RG-L and Dist-2/3, showing that global planning latent variable play an important role in both relevance and diversity of the generated synthetic data. Thus, the combination of  $\mathcal{L}_{\mathrm{global}}$ and $\mathcal{L}_{\mathrm{local}}$ guarantees the quality of generated synthetic dialogues.

\begin{table}[!t]
    \centering
    \tiny
    \begin{tabular}{l}
    \toprule
        \colorbox{blue!10}{\textbf{Covered Flowchart paths}}: car won't start $\rightarrow$ starter crank? Yes $\rightarrow$ Engine fires? No \\ $\rightarrow$ Spark to plugs? No  $\rightarrow$ Spark from coil? No  $\rightarrow$ 12V+ at coil primary? No \\ $\rightarrow$ Ignition system wiring, voltage regulator. \\ \hline
        \colorbox{black!5}{\textbf{Gen-1:}} \\ \textbf{user:} My family and children were very happy for the week end trip in my Honda CR-V. \\When everything was packed and we got seated the car would not simply start. \textbf{(statement)}\\
 \textbf{agent:} After the key is turned does the starter crank? \textbf{(yes-no-q)} \\
 \textbf{user:} Yes, The starter cranks when the key is turned \textbf{(inform)} \\
 \textbf{agent:} When you turn the key, does the engine fire right up? \textbf{(yes-no-q)} \\
 \textbf{user:} No, When the key is turned the engine does not fire up. \textbf{(inform)} \\
 \textbf{agent:} Are the sparks able to reach the plug? \textbf{(yes-no-q)} \\
 \textbf{user:} How do I check if it is able to reach the plug?  \textbf{( \colorbox{yellow!15}{clarification})} \\
 \textbf{agent:} Hold the metal clip of the spark plug wire with a screw driver after pulling it and \\ ask your assistant to crank it over. A bright spark will be visible to you. \textbf{(\colorbox{yellow!15}{suggestion})} \\
 \textbf{user:}No, I'm not seeing any evidence of that. The sparks aren't reaching the plug. \textbf{(inform)}  \\
\textbf{agent:}  Did spark come out of the ignition coil? \textbf{(yes-no-q)} \\
 \textbf{user:}There is no spark near the ignition coil. \textbf{(inform)} \\
 \textbf{agent:} Does the voltmeter reading show 12 volt+ at the primary coil?  \textbf{(yes-no-q)}\\
 \textbf{user:}No, the voltmeter reading does not show 12 volt + at the primary coil.  \textbf{(inform)}\\
 \textbf{agent:} There are a couple of suggestions.  Either your car ignition system wiring or the \\voltage regulator is at fault. \textbf{(suggestion)} \\ \hline \hline
  \colorbox{black!5}{\textbf{Gen-2:}} \\
\textbf{user:} I am unable to start my Lincoln Continental. I think the problem is major, because \\I've only had the car for a week. \textbf{(statement)}\\
 \textbf{agent:}Does turning the key make the starter crank? \textbf{(yes-no-q)} \\
\textbf{user:} It cranks when I turn the key. \textbf{(inform)}\\
\textbf{agent:} Can you tell me if the engine fires on turning the key? \textbf{(yes-no-q)}\\
\textbf{user:} The engine doesn't fire on turning the key. \textbf{(inform)}\\
 \textbf{agent:}Is the spark reaching the plugs? \textbf{(yes-no-q)}\\
\textbf{user:} Spark isn't reaching the plugs. \textbf{(inform)}\\
\textbf{agent:} Is there any kind of spark from the ignition coil? \textbf{(yes-no-q)}\\
\textbf{user:} It's not coming from there. \textbf{(inform)}\\
\textbf{agent:} Is the reading more than 12 volt at the primary coil? \textbf{(yes-no-q)}\\
\textbf{user:} No, the reading is not more than 12 volt at the primary coil. \textbf{(inform)}\\ 
\textbf{agent:}  The problem is the car ignition system wiring or voltage regulator is not working.  \\ Hope I was helpful. \textbf{(suggestion)} \\ \hline\hline
        \colorbox{red!10}{\textbf{Uncovered Flowchart paths}}: laptop wireless problem. $\rightarrow$ See Wireless Network? No $\rightarrow$ \\ Switch on, enabled? Yes  $\rightarrow$ outer active, in range? No  $\rightarrow$ Get router wireless activity, \\ LED blinking, try laptop in same room \\ \hline
        \textbf{user:} I am unable to connect to the wireless network with my Lenovo. Any solution \\ I can fix this? \textbf{(statement)}\\
 \textbf{agent:} Can you see the wireless network you want to connect to? \textbf{(yes-no-q)} \\
 \textbf{user:} How do I check if my laptop to see the wireless network? \textbf{(clarification)} \\
 \textbf{agent:} Click on the network icon, then you can see it. \textbf{(Suggestion)} \\
 \textbf{user:} Sorry, man, it isn't showing up. \textbf{(inform)} \\
 \textbf{agent:} Ok, Is the wireless adapter switched ON?  \textbf{(yes-no-q)} \\
 \textbf{user:} Yes, of course. It's switched on.  \textbf{(Inform)} \\
 \textbf{agent:} Is the router in range and active? \textbf{(yes-no-q)} \\
 \textbf{user:}No, I checked my network router and it's within range and working. \textbf{(inform)}  \\
\textbf{agent:} Make sure the router is turned ON and its wireless activity LED is blinking. \\ You should be able to connect in this way. \textbf{(suggestion)} \\ \hline \hline
    \end{tabular}
    \caption{Pseudo dialogue generation case by our \da model, including {covered path} and {uncovered path}. }
    \label{CS:example}
\end{table}

\subsection{Case Study}
In this section, we conduct a case study towards the output of \da when given covered path and uncovered path as inputs, respectively. The output examples are given in Table~\ref{CS:example}.

\paragraph{\colorbox{blue!10}{Covered path}} 
We first focus on the \da outputs (Gen-1 and Gen-2) for covered flowchart path. By interpolating the latent variables, \da is able to generate diverse utterances along with the corresponding dialogue act sequences. For instance, in addition to generating a series of "yes-no-question" dialogue acts during the problem diagnosis process, \da  can incorporate other dialogue acts as well, such as "\colorbox{yellow!15}{clarification}" and "\colorbox{yellow!15}{suggestion}". This will further lead to the diversity in their corresponding utterance generation. In addition, although being diverse, both example dialogues still maintain high faithfulness towards the input flowchart path.

\paragraph{\colorbox{red!10}{Uncovered Paths}}
As only 65\% flowchart paths are covered in the \textit{FloDial} training data, we conduct a further qualitative analysis  to explore whether \da can generate acceptable synthetic dialogues for those \emph{uncovered} paths.
As shown in the bottom case in Table~\ref{CS:example}, 
we can tell from the example that basic requirements such as fluency, naturalness, and faithfulness have been fulfilled.  
%
We hypothesise that, through fine-tuning on those covered dialogue instances,  dialogue systems trained on \da augmented data acquire and memorize  relevant domain knowledge in flowcharts.
Therefore, these dialogue systems will likely to have better performance compared to the ones which have not seen training data instances for the uncovered flowchart paths. 


\subsection{Human Evaluation}
We have shown that our proposed \da method can achieve better performance in both extrinsic and intrinsic evaluations. However, the automatic metrics do not necessarily reflect human preference of the generated text. We therefore select 150 output samples for each baseline synthetic models and \da model. For each individual sample, we ask three annotators to judge from three aspects: \textit{Faithfulness}, \textit{Relevance} and \textit{Informativeness}. The scale ranges from 0 (low) to 3 (high). Table~\ref{tab_human} summarizes human evaluation results.  The kappa scores indicate that the annotators came to a fair agreement in the judgement. Compared to baseline models, our \da approach achieves higher performance on its generated synthetic dialogues. Thus, synthetic data from \da also aligns well with human preferences.


\begin{table}[!t]
    \centering
    \tiny
    \begin{tabular}{lcccc}
    \toprule
       Model  &  Faith. & Rel. & Info. & \textit{kappa} \\ \hline
       EDA  & {1.37}	& 1.85	& 2.09 & 0.64 \\
       Back-Tran  &  1.62 & 2.27 & 2.18 & 0.59   \\
       GPT-2  &  2.24 & 	{2.53} & 	\textbf{2.65} &  0.56   \\
        BART  & 2.19 & 2.59 & 2.16 &   0.59  \\
      \da & \textbf{2.33} & \textbf{2.60} & 2.54 &  0.57 \\
    \bottomrule
    \end{tabular}
    \caption{Human Evaluation. Annotators are required to judge each instances individually generated by baselines and our model.}
    \label{tab_human}
\end{table}

\section{Related Work}

\subsection{Troubleshooting Dialogue Systems}
Troubleshooting dialogues typically appear in problem-solving scenarios between a novice and an expert~\cite{boye2007dialogue,williams2007applying,janarthanam2008user}. In such scenarios, experts with domain knowledge help novices by asking a series of questions to identify the problem, while the novice mostly supplies answers. 
Recently, \citet{wei2018task} built an end-to-end system for patient diagnosis, and a flowchart-grounded troubleshooting dialogue scenario was proposed by \citep{flowchart2021emnlp}. However, these methods are only explored in limited domains and datasets (\eg computer, car), while \da is a general approach to synthesize pseudo dialogues.


\subsection{Data Augmentation for Dialogue}
Data augmentation
for dialogue-related tasks has been explored in several previous works: \citet{quan2019effective} presented sentence and word-level data augmentation approaches for end-to-end task-oriented dialogues; 
\citet{hou2018sequence} presented a seq2seq framework to augment dialogue utterances for {dialogue language understanding,} including a ranking system to produce diverse utterances; \citet{zhang2020seqmix} proposed a {Multi-Action Data Augmentation (MADA)} model,
which uses dialog states to summarize the dialog history, and then maps  dialog states to their system actions.
Data augmentation methods for spoken dialogue and language understanding, including generative latent variable models, were investigated in~\cite{hou2018sequence,kim2019data,yoo2019data}. {However, most of the previous works focus on data augmentation for discriminative tasks.
\citet{kann2022open} used retrieval-based data augmentation to improve response generation performance in open-domain dialogues, which heavily rely on relevant external resource. Given the limited relevant external resource in FTD, the retrieval-based data augmentation method cannot be applied for FTD systems.

\subsection{Variational Models in Text Generation}
{In addition to data augmentation~\cite{wu2019data,norouzi2020exemplar},}
Variational Autoencoders (VAEs)~\cite{kingma2014auto} are widely used in various text generation tasks, including machine translation~\cite{su2018variational}, question answering~\cite{tang2021continuous}, and dialogue response generation~\cite{serban2017hierarchical,shen2019modeling,zhan2021colv}. {In contrast to previous work using VAE models for data augmentation,}
we devised a model for pseudo dialogue generation that incorporates dialogue features, such as dialogue act and flowchart instruction.

\section{Conclusions}
In this paper, we explore the synthetic dialogue generation as a data augmentation approach with pre-trained model for flowchart-grounded troubleshooting dialogue systems. In further, in order to incorporate dialogue-specific features efficiently, we present a planning-based generative model \da for generating synthetic dialogues on troubleshooting dialogue task. 
The generated augmented dataset is then used to train an FTD systems. Experiments on the \textit{FloDial} benchmark show the effectiveness of our proposed method. In the future, we plan to generalise our method to more complex dialogues, and apply it to other tasks.


\section*{Ethics Statement}

We emphasize several ethical consideration in this work. First, we would like to achknowledge the efforts of crowd-workers and annotators throughout the  dataset annotation and human evaluation processes. This study underwent a thorough review and received approval from an internal board. Every annotator received a compensation of 25 AUD per hour during the annotation and evaluation stages. The associated dataset is strictly for research purpose only.

\bibliography{reference,custom}

\begin{thebibliography}{42}
\expandafter\ifx\csname natexlab\endcsname\relax\def\natexlab#1{#1}\fi

\bibitem[{Benjamini and Hochberg(1995)}]{benjamini1995controlling}
Yoav Benjamini and Yosef Hochberg. 1995.
\newblock Controlling the false discovery rate: a practical and powerful approach to multiple testing.
\newblock \emph{Journal of the Royal statistical society: series B (Methodological)}, 57(1):289--300.

\bibitem[{Boye(2007)}]{boye2007dialogue}
Johan Boye. 2007.
\newblock \href {https://aclanthology.org/2007.sigdial-1.45} {Dialogue management for automatic troubleshooting and other problem-solving applications}.
\newblock In \emph{Proceedings of the 8th SIGdial Workshop on Discourse and Dialogue}, pages 247--255, Antwerp, Belgium. Association for Computational Linguistics.

\bibitem[{Budzianowski et~al.(2018)Budzianowski, Wen, Tseng, Casanueva, Ultes, Ramadan, and Ga{\v{s}}i{\'c}}]{budzianowski2018multiwoz}
Pawe{\l} Budzianowski, Tsung-Hsien Wen, Bo-Hsiang Tseng, I{\~n}igo Casanueva, Stefan Ultes, Osman Ramadan, and Milica Ga{\v{s}}i{\'c}. 2018.
\newblock \href {https://doi.org/10.18653/v1/D18-1547} {{M}ulti{WOZ} - a large-scale multi-domain {W}izard-of-{O}z dataset for task-oriented dialogue modelling}.
\newblock In \emph{Proceedings of the 2018 Conference on Empirical Methods in Natural Language Processing}, pages 5016--5026, Brussels, Belgium. Association for Computational Linguistics.

\bibitem[{Bunt(2011)}]{bunt2011semantics}
Harry Bunt. 2011.
\newblock \href {https://aclanthology.org/W11-0101} {The semantics of dialogue acts}.
\newblock In \emph{Proceedings of the Ninth International Conference on Computational Semantics ({IWCS} 2011)}.

\bibitem[{Dai et~al.(2022)Dai, Chaganty, Zhao, Amini, Rashid, Green, and Guu}]{dai2022dialoginpainting}
Zhuyun Dai, Arun~Tejasvi Chaganty, Vincent~Y Zhao, Aida Amini, Qazi~Mamunur Rashid, Mike Green, and Kelvin Guu. 2022.
\newblock Dialog inpainting: Turning documents into dialogs.
\newblock In \emph{International Conference on Machine Learning}, pages 4558--4586. PMLR.

\bibitem[{Hou et~al.(2018)Hou, Liu, Che, and Liu}]{hou2018sequence}
Yutai Hou, Yijia Liu, Wanxiang Che, and Ting Liu. 2018.
\newblock \href {https://aclanthology.org/C18-1105} {Sequence-to-sequence data augmentation for dialogue language understanding}.
\newblock In \emph{Proceedings of the 27th International Conference on Computational Linguistics}, pages 1234--1245, Santa Fe, New Mexico, USA. Association for Computational Linguistics.

\bibitem[{Janarthanam and Lemon(2008)}]{janarthanam2008user}
Srinivasan Janarthanam and Oliver Lemon. 2008.
\newblock User simulations for online adaptation and knowledge-alignment in troubleshooting dialogue systems.
\newblock \emph{Semantics and Pragmatics of Dialogue (LONDIAL)}, page~45.

\bibitem[{Kann et~al.(2022)Kann, Ebrahimi, Koh, Dudy, and Roncone}]{kann2022open}
Katharina Kann, Abteen Ebrahimi, Joewie Koh, Shiran Dudy, and Alessandro Roncone. 2022.
\newblock \href {https://doi.org/10.18653/v1/2022.nlp4convai-1.13} {Open-domain dialogue generation: What we can do, cannot do, and should do next}.
\newblock In \emph{Proceedings of the 4th Workshop on NLP for Conversational AI}, pages 148--165, Dublin, Ireland. Association for Computational Linguistics.

\bibitem[{Kim et~al.(2019)Kim, Roh, and Kim}]{kim2019data}
Hwa-Yeon Kim, Yoon-Hyung Roh, and Young-Kil Kim. 2019.
\newblock \href {https://doi.org/10.18653/v1/N19-3014} {Data augmentation by data noising for open-vocabulary slots in spoken language understanding}.
\newblock In \emph{Proceedings of the 2019 Conference of the North {A}merican Chapter of the Association for Computational Linguistics: Student Research Workshop}, pages 97--102, Minneapolis, Minnesota. Association for Computational Linguistics.

\bibitem[{Kingma and Welling(2014)}]{kingma2014auto}
Diederik~P. Kingma and Max Welling. 2014.
\newblock \href {http://arxiv.org/abs/1312.6114} {Auto-encoding variational bayes}.
\newblock In \emph{2nd International Conference on Learning Representations, {ICLR} 2014, Banff, AB, Canada, April 14-16, 2014, Conference Track Proceedings}.

\bibitem[{Kingma et~al.(2016)Kingma, Salimans, Jozefowicz, Chen, Sutskever, and Welling}]{kingma2016improved}
Durk~P Kingma, Tim Salimans, Rafal Jozefowicz, Xi~Chen, Ilya Sutskever, and Max Welling. 2016.
\newblock Improved variational inference with inverse autoregressive flow.
\newblock \emph{Advances in neural information processing systems}, 29.

\bibitem[{Kullback and Leibler(1951)}]{kullback1951information}
Solomon Kullback and Richard~A Leibler. 1951.
\newblock On information and sufficiency.
\newblock \emph{The annals of mathematical statistics}, 22(1):79--86.

\bibitem[{Leake et~al.(2005)Leake, Bogaerts, Evans, McMullen, Oder, and Valerio}]{leake2005using}
David~B Leake, Steven Bogaerts, Michael Evans, Rick McMullen, Michael Oder, and Alejandro Valerio. 2005.
\newblock Using cases to support divergent roles in distributed collaboration.
\newblock In \emph{FLAIRS Conference}, pages 117--122.

\bibitem[{Lewis et~al.(2020)Lewis, Liu, Goyal, Ghazvininejad, Mohamed, Levy, Stoyanov, and Zettlemoyer}]{lewis2019bart}
Mike Lewis, Yinhan Liu, Naman Goyal, Marjan Ghazvininejad, Abdelrahman Mohamed, Omer Levy, Veselin Stoyanov, and Luke Zettlemoyer. 2020.
\newblock \href {https://doi.org/10.18653/v1/2020.acl-main.703} {{BART}: Denoising sequence-to-sequence pre-training for natural language generation, translation, and comprehension}.
\newblock In \emph{Proceedings of the 58th Annual Meeting of the Association for Computational Linguistics}, pages 7871--7880, Online. Association for Computational Linguistics.

\bibitem[{Li et~al.(2016)Li, Galley, Brockett, Gao, and Dolan}]{li2016distinct}
Jiwei Li, Michel Galley, Chris Brockett, Jianfeng Gao, and Bill Dolan. 2016.
\newblock \href {https://doi.org/10.18653/v1/N16-1014} {A diversity-promoting objective function for neural conversation models}.
\newblock In \emph{Proceedings of the 2016 Conference of the North {A}merican Chapter of the Association for Computational Linguistics: Human Language Technologies}, pages 110--119, San Diego, California. Association for Computational Linguistics.

\bibitem[{Li and Liang(2021)}]{li-liang-2021-prefix}
Xiang~Lisa Li and Percy Liang. 2021.
\newblock \href {https://doi.org/10.18653/v1/2021.acl-long.353} {Prefix-tuning: Optimizing continuous prompts for generation}.
\newblock In \emph{Proceedings of the 59th Annual Meeting of the Association for Computational Linguistics and the 11th International Joint Conference on Natural Language Processing (Volume 1: Long Papers)}, pages 4582--4597, Online. Association for Computational Linguistics.

\bibitem[{Lin(2004)}]{lin2004rouge}
Chin-Yew Lin. 2004.
\newblock \href {https://aclanthology.org/W04-1013} {{ROUGE}: A package for automatic evaluation of summaries}.
\newblock In \emph{Text Summarization Branches Out}, pages 74--81, Barcelona, Spain. Association for Computational Linguistics.

\bibitem[{Loshchilov and Hutter(2019)}]{loshchilov2017decoupled}
Ilya Loshchilov and Frank Hutter. 2019.
\newblock \href {https://openreview.net/forum?id=Bkg6RiCqY7} {Decoupled weight decay regularization}.
\newblock In \emph{7th International Conference on Learning Representations, {ICLR} 2019, New Orleans, LA, USA, May 6-9, 2019}. OpenReview.net.

\bibitem[{Norouzi et~al.(2020)Norouzi, Fleet, and Norouzi}]{norouzi2020exemplar}
Sajad Norouzi, David~J. Fleet, and Mohammad Norouzi. 2020.
\newblock \href {https://proceedings.neurips.cc/paper/2020/hash/63c17d596f401acb520efe4a2a7a01ee-Abstract.html} {Exemplar {VAE:} linking generative models, nearest neighbor retrieval, and data augmentation}.
\newblock In \emph{Advances in Neural Information Processing Systems 33: Annual Conference on Neural Information Processing Systems 2020, NeurIPS 2020, December 6-12, 2020, virtual}.

\bibitem[{Paek and Pieraccini(2008)}]{paek2008automating}
Tim Paek and Roberto Pieraccini. 2008.
\newblock Automating spoken dialogue management design using machine learning: An industry perspective.
\newblock \emph{Speech communication}, pages 716--729.

\bibitem[{Papineni et~al.(2002)Papineni, Roukos, Ward, and Zhu}]{papineni2002bleu}
Kishore Papineni, Salim Roukos, Todd Ward, and Wei-Jing Zhu. 2002.
\newblock \href {https://doi.org/10.3115/1073083.1073135} {{B}leu: a method for automatic evaluation of machine translation}.
\newblock In \emph{Proceedings of the 40th Annual Meeting of the Association for Computational Linguistics}, pages 311--318, Philadelphia, Pennsylvania, USA. Association for Computational Linguistics.

\bibitem[{Quan and Xiong(2019)}]{quan2019effective}
Jun Quan and Deyi Xiong. 2019.
\newblock Effective data augmentation approaches to end-to-end task-oriented dialogue.
\newblock In \emph{2019 International Conference on Asian Language Processing (IALP)}, pages 47--52. IEEE.

\bibitem[{Radford et~al.(2019)Radford, Wu, Child, Luan, Amodei, Sutskever et~al.}]{radford2019language}
Alec Radford, Jeffrey Wu, Rewon Child, David Luan, Dario Amodei, Ilya Sutskever, et~al. 2019.
\newblock Language models are unsupervised multitask learners.
\newblock \emph{OpenAI blog}, 1(8):9.

\bibitem[{Raghu et~al.(2021)Raghu, Agarwal, Joshi, and {Mausam}}]{flowchart2021emnlp}
Dinesh Raghu, Shantanu Agarwal, Sachindra Joshi, and {Mausam}. 2021.
\newblock \href {https://doi.org/10.18653/v1/2021.emnlp-main.357} {End-to-end learning of flowchart grounded task-oriented dialogs}.
\newblock In \emph{Proceedings of the 2021 Conference on Empirical Methods in Natural Language Processing}, pages 4348--4366, Online and Punta Cana, Dominican Republic. Association for Computational Linguistics.

\bibitem[{Roller et~al.(2021)Roller, Dinan, Goyal, Ju, Williamson, Liu, Xu, Ott, Smith, Boureau, and Weston}]{roller2020recipes}
Stephen Roller, Emily Dinan, Naman Goyal, Da~Ju, Mary Williamson, Yinhan Liu, Jing Xu, Myle Ott, Eric~Michael Smith, Y-Lan Boureau, and Jason Weston. 2021.
\newblock \href {https://doi.org/10.18653/v1/2021.eacl-main.24} {Recipes for building an open-domain chatbot}.
\newblock In \emph{Proceedings of the 16th Conference of the European Chapter of the Association for Computational Linguistics: Main Volume}, pages 300--325, Online. Association for Computational Linguistics.

\bibitem[{Sennrich et~al.(2016)Sennrich, Haddow, and Birch}]{sennrich2016improving}
Rico Sennrich, Barry Haddow, and Alexandra Birch. 2016.
\newblock \href {https://doi.org/10.18653/v1/P16-1009} {Improving neural machine translation models with monolingual data}.
\newblock In \emph{Proceedings of the 54th Annual Meeting of the Association for Computational Linguistics (Volume 1: Long Papers)}, pages 86--96, Berlin, Germany. Association for Computational Linguistics.

\bibitem[{Serban et~al.(2017)Serban, Sordoni, Lowe, Charlin, Pineau, Courville, and Bengio}]{serban2017hierarchical}
Iulian~Vlad Serban, Alessandro Sordoni, Ryan Lowe, Laurent Charlin, Joelle Pineau, Aaron~C. Courville, and Yoshua Bengio. 2017.
\newblock \href {http://aaai.org/ocs/index.php/AAAI/AAAI17/paper/view/14567} {A hierarchical latent variable encoder-decoder model for generating dialogues}.
\newblock In \emph{Proceedings of the Thirty-First {AAAI} Conference on Artificial Intelligence, February 4-9, 2017, San Francisco, California, {USA}}, pages 3295--3301. {AAAI} Press.

\bibitem[{Shen et~al.(2019)Shen, Feng, and Zhan}]{shen2019modeling}
Lei Shen, Yang Feng, and Haolan Zhan. 2019.
\newblock Modeling semantic relationship in multi-turn conversations with hierarchical latent variables.
\newblock \emph{arXiv preprint arXiv:1906.07429}.

\bibitem[{Stolcke et~al.(2000)Stolcke, Ries, Coccaro, Shriberg, Bates, Jurafsky, Taylor, Martin, Van Ess-Dykema, and Meteer}]{stolcke2000dialogue}
Andreas Stolcke, Klaus Ries, Noah Coccaro, Elizabeth Shriberg, Rebecca Bates, Daniel Jurafsky, Paul Taylor, Rachel Martin, Carol Van Ess-Dykema, and Marie Meteer. 2000.
\newblock \href {https://aclanthology.org/J00-3003} {Dialogue act modeling for automatic tagging and recognition of conversational speech}.
\newblock \emph{Computational Linguistics}, 26(3):339--374.

\bibitem[{Su et~al.(2018)Su, Wu, Xiong, Lu, Han, and Zhang}]{su2018variational}
Jinsong Su, Shan Wu, Deyi Xiong, Yaojie Lu, Xianpei Han, and Biao Zhang. 2018.
\newblock \href {https://www.aaai.org/ocs/index.php/AAAI/AAAI18/paper/view/16791} {Variational recurrent neural machine translation}.
\newblock In \emph{Proceedings of the Thirty-Second {AAAI} Conference on Artificial Intelligence, (AAAI-18), the 30th innovative Applications of Artificial Intelligence (IAAI-18), and the 8th {AAAI} Symposium on Educational Advances in Artificial Intelligence (EAAI-18), New Orleans, Louisiana, USA, February 2-7, 2018}, pages 5488--5495. {AAAI} Press.

\bibitem[{Tang et~al.(2021)Tang, Zhang, Kim, and Bansal}]{tang2021continuous}
Zineng Tang, Shiyue Zhang, Hyounghun Kim, and Mohit Bansal. 2021.
\newblock \href {https://doi.org/10.18653/v1/2021.acl-long.355} {Continuous language generative flow}.
\newblock In \emph{Proceedings of the 59th Annual Meeting of the Association for Computational Linguistics and the 11th International Joint Conference on Natural Language Processing (Volume 1: Long Papers)}, pages 4609--4622, Online. Association for Computational Linguistics.

\bibitem[{Wei and Zou(2019)}]{wei-zou-2019-eda}
Jason Wei and Kai Zou. 2019.
\newblock \href {https://doi.org/10.18653/v1/D19-1670} {{EDA}: Easy data augmentation techniques for boosting performance on text classification tasks}.
\newblock In \emph{Proceedings of the 2019 Conference on Empirical Methods in Natural Language Processing and the 9th International Joint Conference on Natural Language Processing (EMNLP-IJCNLP)}, pages 6382--6388, Hong Kong, China. Association for Computational Linguistics.

\bibitem[{Wei et~al.(2018)Wei, Liu, Peng, Tou, Chen, Huang, Wong, and Dai}]{wei2018task}
Zhongyu Wei, Qianlong Liu, Baolin Peng, Huaixiao Tou, Ting Chen, Xuanjing Huang, Kam-fai Wong, and Xiangying Dai. 2018.
\newblock \href {https://doi.org/10.18653/v1/P18-2033} {Task-oriented dialogue system for automatic diagnosis}.
\newblock In \emph{Proceedings of the 56th Annual Meeting of the Association for Computational Linguistics (Volume 2: Short Papers)}, pages 201--207, Melbourne, Australia. Association for Computational Linguistics.

\bibitem[{Wen et~al.(2017)Wen, Vandyke, Mrk{\v{s}}i{\'c}, Ga{\v{s}}i{\'c}, Rojas-Barahona, Su, Ultes, and Young}]{wen2017network}
Tsung-Hsien Wen, David Vandyke, Nikola Mrk{\v{s}}i{\'c}, Milica Ga{\v{s}}i{\'c}, Lina~M. Rojas-Barahona, Pei-Hao Su, Stefan Ultes, and Steve Young. 2017.
\newblock \href {https://aclanthology.org/E17-1042} {A network-based end-to-end trainable task-oriented dialogue system}.
\newblock In \emph{Proceedings of the 15th Conference of the {E}uropean Chapter of the Association for Computational Linguistics: Volume 1, Long Papers}, pages 438--449, Valencia, Spain. Association for Computational Linguistics.

\bibitem[{Williams(2007)}]{williams2007applying}
Jason Williams. 2007.
\newblock \href {https://aclanthology.org/W07-0301} {Applying {POMDP}s to dialog systems in the troubleshooting domain}.
\newblock In \emph{Proceedings of the Workshop on Bridging the Gap: Academic and Industrial Research in Dialog Technologies}, pages 1--8, Rochester, NY. Association for Computational Linguistics.

\bibitem[{Wu et~al.(2019)Wu, Wang, Qian, and Yu}]{wu2019data}
Zhanghao Wu, Shuai Wang, Yanmin Qian, and Kai Yu. 2019.
\newblock \href {https://doi.org/10.21437/Interspeech.2019-2248} {Data augmentation using variational autoencoder for embedding based speaker verification}.
\newblock In \emph{Interspeech 2019, 20th Annual Conference of the International Speech Communication Association, Graz, Austria, 15-19 September 2019}, pages 1163--1167. {ISCA}.

\bibitem[{Yoo et~al.(2019)Yoo, Shin, and Lee}]{yoo2019data}
Kang~Min Yoo, Youhyun Shin, and Sang{-}goo Lee. 2019.
\newblock \href {https://doi.org/10.1609/aaai.v33i01.33017402} {Data augmentation for spoken language understanding via joint variational generation}.
\newblock In \emph{The Thirty-Third {AAAI} Conference on Artificial Intelligence, {AAAI} 2019, The Thirty-First Innovative Applications of Artificial Intelligence Conference, {IAAI} 2019, The Ninth {AAAI} Symposium on Educational Advances in Artificial Intelligence, {EAAI} 2019, Honolulu, Hawaii, USA, January 27 - February 1, 2019}, pages 7402--7409. {AAAI} Press.

\bibitem[{Yuan et~al.(2021)Yuan, Neubig, and Liu}]{yuan2021bartscore}
Weizhe Yuan, Graham Neubig, and Pengfei Liu. 2021.
\newblock Bartscore: Evaluating generated text as text generation.
\newblock \emph{Advances in Neural Information Processing Systems}, 34:27263--27277.

\bibitem[{Zhan et~al.(2021)Zhan, Shen, Chen, and Zhang}]{zhan2021colv}
Haolan Zhan, Lei Shen, Hongshen Chen, and Hainan Zhang. 2021.
\newblock Colv: A collaborative latent variable model for knowledge-grounded dialogue generation.
\newblock In \emph{Proceedings of the 2021 Conference on Empirical Methods in Natural Language Processing}, pages 2250--2261.

\bibitem[{Zhang et~al.(2020{\natexlab{a}})Zhang, Ou, and Yu}]{zhang2020seqmix}
Yichi Zhang, Zhijian Ou, and Zhou Yu. 2020{\natexlab{a}}.
\newblock \href {https://aaai.org/ojs/index.php/AAAI/article/view/6507} {Task-oriented dialog systems that consider multiple appropriate responses under the same context}.
\newblock In \emph{The Thirty-Fourth {AAAI} Conference on Artificial Intelligence, {AAAI} 2020, The Thirty-Second Innovative Applications of Artificial Intelligence Conference, {IAAI} 2020, The Tenth {AAAI} Symposium on Educational Advances in Artificial Intelligence, {EAAI} 2020, New York, NY, USA, February 7-12, 2020}, pages 9604--9611. {AAAI} Press.

\bibitem[{Zhang et~al.(2020{\natexlab{b}})Zhang, Sun, Galley, Chen, Brockett, Gao, Gao, Liu, and Dolan}]{zhang2019dialogpt}
Yizhe Zhang, Siqi Sun, Michel Galley, Yen-Chun Chen, Chris Brockett, Xiang Gao, Jianfeng Gao, Jingjing Liu, and Bill Dolan. 2020{\natexlab{b}}.
\newblock \href {https://doi.org/10.18653/v1/2020.acl-demos.30} {{DIALOGPT} : Large-scale generative pre-training for conversational response generation}.
\newblock In \emph{Proceedings of the 58th Annual Meeting of the Association for Computational Linguistics: System Demonstrations}, pages 270--278, Online. Association for Computational Linguistics.

\bibitem[{Zhu et~al.(2018)Zhu, Lu, Zheng, Guo, Zhang, Wang, and Yu}]{zhu2018selfbleu}
Yaoming Zhu, Sidi Lu, Lei Zheng, Jiaxian Guo, Weinan Zhang, Jun Wang, and Yong Yu. 2018.
\newblock \href {https://doi.org/10.1145/3209978.3210080} {Texygen: {A} benchmarking platform for text generation models}.
\newblock In \emph{The 41st International {ACM} {SIGIR} Conference on Research {\&} Development in Information Retrieval, {SIGIR} 2018, Ann Arbor, MI, USA, July 08-12, 2018}, pages 1097--1100. {ACM}.

\end{thebibliography}
\bibliographystyle{acl_natbib}

\newpage
\newpage
\appendix

\section{Appendix}


\subsection{Derivation of Variational Lower Bound}

\begin{align*}
    &\text{log}p_{\theta}({\bf a, y}|{\bf x}) \\
    &=\text{log}\int_{\textbf{z}_{a}}\int_{\textbf{z}_{y}}p_{\theta}({\bf a}|\textbf{z}_{a}, {\bf x})\cdot \\
    &p_{\theta}({\bf y}|\textbf{z}_{y}, \textbf{a}, {\bf x})p_{\phi}(\textbf{z}_{y}|\textbf{a},\textbf{x})p_{\phi}(\textbf{z}_{a}|\textbf{x})d_{\textbf{z}_{a}} \\
    &=\text{log}\int_{\textbf{z}_{a}}p_{\theta}({\bf a}|\textbf{z}_{a}, {\bf x})p_{\phi}(\textbf{z}_{a}|\textbf{x})\frac{q_{\phi}(\textbf{z}_{a} | \textbf{x},\textbf{y})}{q_{\phi}(\textbf{z}_{a} | \textbf{x},\textbf{y})}\cdot \\
    &\int_{\textbf{z}_{y}}p_{\theta}({\bf y}|\textbf{z}_{y}, \textbf{a}, {\bf x})p_{\phi}(\textbf{z}_{y}|\textbf{a},\textbf{x})\frac{q_{\phi}(\textbf{z}_{y} | \textbf{x},\textbf{a}, \textbf{y})}{q_{\phi}(\textbf{z}_{y} | \textbf{x},\textbf{a}, \textbf{y})}d_{\textbf{z}_{x}} \\
    &=\text{log}\int_{\textbf{z}_{a}}p_{\theta}({\bf a}|\textbf{z}_{a}, {\bf x})p_{\phi}(\textbf{z}_{a}|\textbf{x})\frac{q_{\phi}(\textbf{z}_{a} | \textbf{x},\textbf{y})}{q_{\phi}(\textbf{z}_{a} | \textbf{x},\textbf{y})}\cdot \\
    &\mathbb{E}_{q_{\phi}(\textbf{z}_{y} | \textbf{x},\textbf{a}, \textbf{y})}\left[ \frac{p_{\theta}({\bf y}|\textbf{z}_{y}, \textbf{a}, {\bf x})p_{\phi}(\textbf{z}_{y}|\textbf{a},\textbf{x})}{q_{\phi}(\textbf{z}_{y} | \textbf{x},\textbf{a}, \textbf{y})} \right]d_{\textbf{z}_{x}} \\
    &=\text{log}\mathbb{E}_{q_{\phi}(\textbf{z}_{a} | \textbf{x},\textbf{y})}\{\frac{p_{\theta}({\bf a}|\textbf{z}_{a}, {\bf x})p_{\phi}(\textbf{z}_{a}|\textbf{x})}{q_{\phi}(\textbf{z}_{a} | \textbf{x},\textbf{y})}\cdot \\
    &\mathbb{E}_{q_{\phi}(\textbf{z}_{y} | \textbf{x},\textbf{a}, \textbf{y})}\left[ \frac{p_{\theta}({\bf y}|\textbf{z}_{y}, \textbf{a}, {\bf x})p_{\phi}(\textbf{z}_{y}|\textbf{a},\textbf{x})}{q_{\phi}(\textbf{z}_{y} | \textbf{x},\textbf{a}, \textbf{y})} \right] \} \\
    &\geq  \mathbb{E}_{q_{\phi}(\textbf{z}_{a} | \textbf{x},\textbf{y})}\{\text{log}\frac{p_{\theta}({\bf a}|\textbf{z}_{a}, {\bf x})p_{\phi}(\textbf{z}_{a}|\textbf{x})}{q_{\phi}(\textbf{z}_{a} | \textbf{x},\textbf{y})} + \\
    &\mathbb{E}_{q_{\phi}(\textbf{z}_{y} | \textbf{x},\textbf{a}, \textbf{y})}\left[ \frac{p_{\theta}({\bf y}|\textbf{z}_{y}, \textbf{a}, {\bf x})p_{\phi}(\textbf{z}_{y}|\textbf{a},\textbf{x})}{q_{\phi}(\textbf{z}_{y} | \textbf{x},\textbf{a}, \textbf{y})} \right] \} \\
    &\approx -KL(q_\phi({\bf z}_{\bf a}|{\bf x}, {\bf y})||p_\theta({\bf z}_{\bf a}|{\bf x})) \\
    & +\mathbb{E}_{{\bf z}_{\bf a} \sim q_\phi}[\log p_{\theta}({a} |{\bf z}_{\bf a},\textbf{x})] \\ &-KL(q_\phi({\bf z}_{\bf y}|{\bf x},{ a},{\bf y})||p_\theta({\bf z}_{\bf y}|{\bf x},{ a})) \\
    & +\mathbb{E}_{{\bf z}_{\bf y} \sim q_\phi}[\log p_{\theta}(\textbf{y} |\mathbf{x}, a, \mathbf{z}_y)]
\end{align*}

\newpage

\subsection{Details about \textit{FloDial} Dataset}

The \textit{FloDial} dataset is collected for the troubleshooting situations, where the interactions between user and agent are carried to diagnose user's problem in specific domain. \textit{FloDial} contains two main domain: vehicle and laptop. Each domain contains 5 sub-problems. For each sub-problem, there is a corresponding flowchart. Dialogues are conducted based on these flowcharts.  Details about each sub-problems and flowchart are shown in Table~\ref{tab:flodial}. \textit{FloDial} contains 1,789 dialogue sessions in total. In the experiments of \textit{FloDial} paper, they construct two settings: \textit{In-Flowchart} and \textit{Out-of-Flowchart} settings. The test set of  \textit{In-Flowchart} setting contains the dialogue in 8 sub-problems (including ticking, brake, battery, wont\_start, drive, overheating, power and lcd), which maintains the same domain with training set. Beside, the test set of \textit{Out-of-Flowchart} setting only contains 2 sub-problems (engine, wireless), while all other 8 sub-problems are treated as training set. An example of flowchart in \textit{car\_wont\_start} domain is shwon in Figure~\ref{ex:flow}

\begin{table}[t]
    \centering
    \scriptsize
    \begin{tabular}{c|ccccc}
    \toprule
       Domain  & \multicolumn{5}{c}{Vehicle}  \\\hline
        ~ & ticking & brake & battery & wont\_start & engine \\\hline
        \#Dialog & 178 & 188 & 196 & 174 & 168 \\\hline
        \#path & 15 & 19 & 18 & 17 & 14 \\\hline\hline
        Domain  & \multicolumn{5}{c}{Laptop}  \\\hline
        ~ & drive & overheating & power & lcd & wireless \\\hline
        \#Dialog & 192 & 186 & 188 & 178 & 196 \\\hline
        \#path & 16 & 13 & 15 & 15 & 15 \\\hline\hline
        
    \end{tabular}
    \caption{\#Dialog and \#sub-path denote the number of dialogue session, and the number of sub-paths of each corresponding flowchart.}
    \label{tab:flodial}
\end{table}

\begin{figure}
    \centering
    \small
    \includegraphics{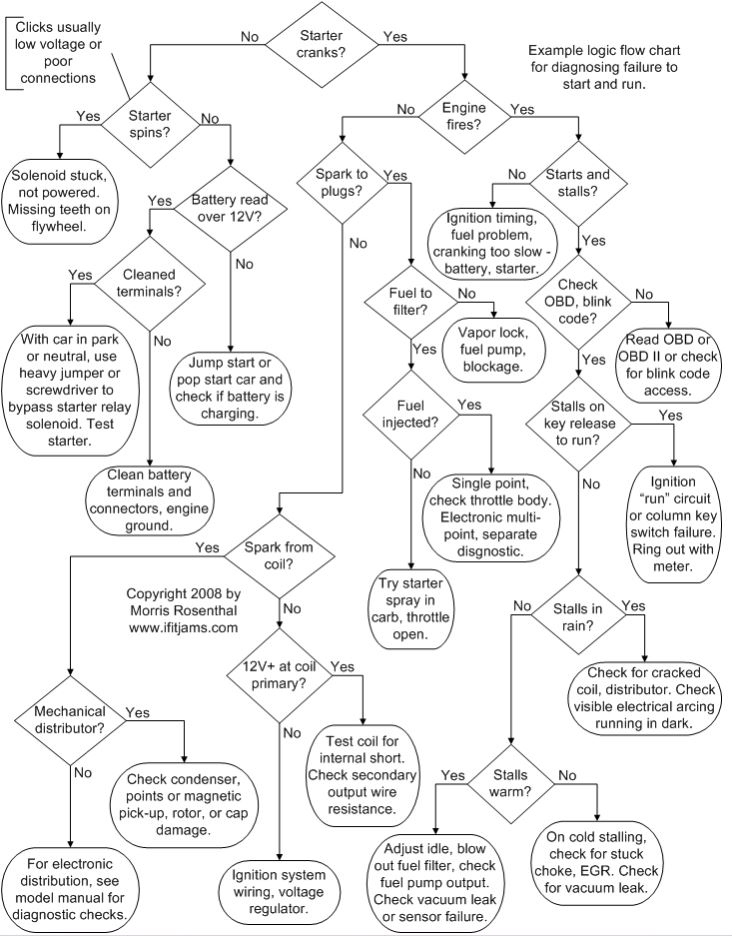}
    \caption{The flowchart example of \textit{car\_wont\_start} domain. The figure is directly downloaded from the website:\url{https://www.ifitjams.com/}, the original source of \textit{FloDial} dataset.}
    \label{ex:flow}
\end{figure}

Besides, as the original \textit{FloDial} dataset does not contain any dialogue act information, we manually label the dialogue act for each dialogue utterance. The selection of dialogue acts is based on the investigation on previous work, including Switchboard (\url{https://catalog.ldc.upenn.edu/LDC97S62}), AMI (\url{https://groups.inf.ed.ac.uk/ami/corpus/}), MultiWoz~\cite{budzianowski2018multiwoz} and etc. 
Finally, we chose seven most frequent dialogue, which also compatible with the \textit{FloDial} dataset. These dialogue acts include: \{statement, inform, yes-no-question, clarification, thanking, closing and suggestion\}. The percentage of each dialogue act in the \textit{FloDial} is: statement: 11.6\%, inform: 34.7\%, yes-no-question: 26.2\%, clarification: 9.8\%, thanking: 6.2\%, closing: 4.3\% and suggestion: 7.2\%.

\label{sec:appendix}

\end{document}